\documentclass{ws-us}
\usepackage[sort,compress,super]{cite}
\begin{document}
\setlength{\pdfpagewidth}{8.5in}
\setlength{\pdfpageheight}{11in}
\catchline{0}{0}{2025}{}{}

\markboth{R.K. Chhetri et al.}{A Survey of Medical Drones from Flight Dynamics, Guidance, Navigation, and Control Perspectives}

\title{A Survey of Medical Drones from Flight Dynamics, Guidance, Navigation, and Control Perspectives}

\newcommand{\myself}{Khamvilai, T}

\author{Roshan Kumar Chhetri$^a$, Sarocha Jetawatthana$^a$, Thanakorn Khamvilai$^{a}$}

\address{$^a$Department of Mechanical Engineering, Texas Tech University, Lubbock, TX 79409, USA\\
Email: roschhet@ttu.edu, sjetawat@ttu.edu, thanakorn.khamvilai@ttu.edu}

\maketitle

\begin{abstract}
The integration of drones into the medical field has revolutionized healthcare delivery by enabling rapid transportation of medical supplies, organs, and even emergency assistance in remote or disaster-stricken areas. While other survey papers focus on the healthcare supply chain, operations, and medical emergency response aspects, this paper provides a comprehensive review of medical drones from the perspectives of flight dynamics and guidance, navigation, and control (GNC) systems. We first discuss the medical aerial delivery mission requirements and suitable uncrewed aerial system (UAS) configurations. We then address payload container design and optimization, and its effect on supplies and overall flight dynamics. We also explore the fundamental principles of GNC in the context of medical drone operations, highlighting key challenges arising from vibration, air temperature, pressure, and humidity, which affect the quality of medical supplies. The paper examines various GNC algorithms that can mitigate these challenges, as well as the algorithms' limitations. With these considerations, this survey aims to provide insights into optimizing GNC frameworks for medical drones, emphasizing research gaps and directions to improve real-world healthcare applications.
\end{abstract}


\begin{multicols}{2}
\section{Introduction}
The integration of drones, or uncrewed aerial systems (UAS), into the medical field has begun to revolutionize healthcare delivery. By enabling the rapid transportation of medical supplies, organs, and even emergency assistance, these systems can overcome geographical barriers and provide critical care in remote or disaster-stricken areas. The primary mission of medical drones is to ensure the timely, reliable, and safe aerial transport of essential supplies, which can include everything from vaccines and medications to blood units and diagnostic samples. The importance of this technology lies in its ability to operate within strict timeframes, which is crucial for preserving the efficacy of time- and temperature-sensitive medical materials.

While many existing recent survey papers focus on the healthcare supply chain, logistics, and emergency response aspects of medical drones \cite{stierlin2024current,roberts2023current,10967690,jadhav2025comprehensive,javaid2025significant}, there is a noticeable gap in literature that provides a comprehensive review from the perspectives of flight dynamics and guidance, navigation, and control (GNC) systems. This paper aims to fill that gap. The purpose of this work is to provide a detailed survey of the core engineering challenges and solutions that define the performance and reliability of medical drone operations. Its significance lies in consolidating the GNC-related knowledge necessary to move this technology from conceptual stages to widespread, real-world application.

This review will carefully examine the current state of the field by first discussing medical aerial delivery mission requirements and the suitable UAS configurations, e.g., fixed-wing, multirotor, and hybrid, to meet them. Subsequently, we address the critical role of payload container design, including fixed, cable-suspended, and temperature-controlled systems, and analyze its effect on both the medical supplies and the overall flight dynamics of the aircraft. We explore the fundamental principles of GNC, highlighting key challenges that arise from environmental factors such as vibration, air temperature, pressure, and humidity, all of which can compromise the quality of medical payloads. This paper also examines various GNC algorithms designed to mitigate these challenges and discusses their limitations. Ultimately, the main aim of this work is to provide critical insights into optimizing GNC frameworks for medical drones, highlighting existing research gaps and proposing future directions to improve the safety, efficiency, and reliability of these life-saving healthcare applications.

\section{Mission Objective and UAS Configurations}
\subsection{Mission Requirements}
The primary mission of medical drones is to enhance healthcare services by ensuring the timely, reliable, and safe aerial transport of essential medical supplies. These supplies may include vaccines, medications, blood units, diagnostic samples, and organs for transplantation. Mission requirements change with each case, but generally include several basic aspects. Time-sensitivity is crucial, as delivery must occur within strict timeframes to preserve the efficacy of medical materials, particularly those requiring cold-chain conditions. The operational range must be sufficient to cover long distances, often over rural or disaster-affected areas with poor infrastructure. High accuracy in precision landing is also important, especially when drones need to deliver packages in confined or predefined areas such as hospital rooftops or remote clinics. Additionally, the UAS must operate with a high degree of autonomy and safety, minimizing human intervention while ensuring reliable performance, especially during emergencies or BVLOS (Beyond Visual Line of Sight) operations. Finally, these systems must adhere to the regulations set by civil aviation authorities, including compliance with communication, navigation, and surveillance protocols and safety standards \cite{lee2022safety}.

\subsection{Medical Delivery UAS}
Medical delivery drones, as shown in Figure \ref{fig:1}, are generally engineered with mission-specific configurations that accommodate payload constraints, environmental robustness, and navigation accuracy. Fixed-wing UAS are often used for long-range missions because of their aerodynamic efficiency and faster cruising speeds. However, their reliance on runways or catapult launches, combined with limited hovering capabilities, makes them less ideal for operations in tight or urban environments \cite{liu2024runway}. On the other hand, multirotor UAVs, like quadcopters and hexacopters, offer vertical takeoff and landing (VTOL), stable control, and the ability to hover in place. These features make them ideal for short-distance deliveries, although their endurance and range are limited compared to fixed-wing UAVs \cite{peksa2024review}. An increasingly popular approach to balance these trade-offs is using hybrid VTOL drones, which integrate the advantages of both fixed-wing and multirotor designs. These systems offer vertical lift for takeoff and landing and efficient forward flight for longer missions, making them ideal for medical deliveries in remote and hard-to-reach regions \cite{osman2025hybrid}. Additionally, some UAVs incorporate modular payload bays equipped with thermal insulation or shock-absorbing components to preserve the biomedical integrity of transported items. Choosing an appropriate UAS configuration ultimately depends on specific mission parameters such as distance, terrain, payload type, and environmental conditions.
\begin{figurehere}
\begin{center}
\centerline{\includegraphics[width=3in]{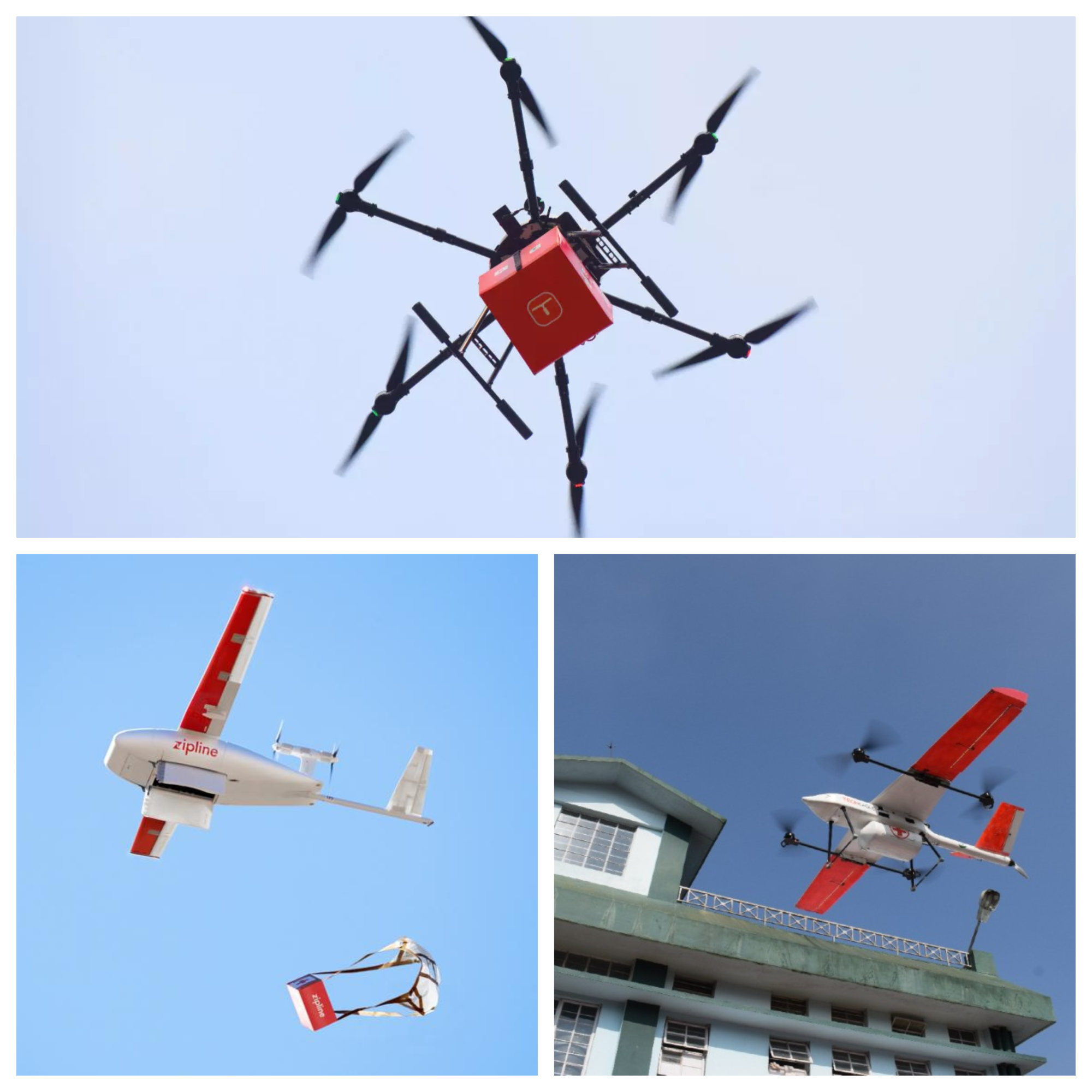}}
\caption{Multi-Rotor \cite{stattimes2025}, Fixed Wing \cite{koetsier2022} and Hybrid UAV Configuration. \cite{stattimes_techeagle2021}}
\label{fig:1}
\end{center}
\end{figurehere}

\subsection{Environmental Effects on Flight Performance}
Medical drone operations are susceptible to environmental conditions, which can significantly impact flight performance and the quality of medical supplies transported. Wind and turbulence can destabilize flight trajectories \cite{doi:10.2514/6.2025-2427}, especially for lightweight drones, making it essential to use adaptive flight control systems to ensure stability in various atmospheric conditions \cite{ma2024deep,axten2023vtol}. Temperature is another critical factor, as low ambient temperature reduces battery capacity, diminishing flight time and operational reliability, while high temperature can cause overheating and shorten battery lifespan \cite{anoune2025performance}. High ambient temperature reduces air density, which lowers available life and requires higher rotor speeds, thereby increasing energy consumption. Also, thermal stress on onboard electronics and motors under temperature extremes can reduce component reliability, force thermal throttling, or even lead to shutdown in worst-case scenarios \cite{pamula2025thermal}. In extreme heat, battery discharge efficiency degrades and internal resistance increases, which results in shorter flight times. Conversely, very cold temperatures slow the chemical reactions inside LiPo batteries, reducing their effective capacity and limiting power delivery to motors \cite{ercan2025effect}.   It was found that the air temperature is one of the meteorological factors that constrain drone endurance and control stability in real flight environments \cite{gao2021weather}.  Simultaneously, payload containers must maintain internal thermal stability to protect perishable or temperature-sensitive materials \cite{waters2024design}. Humidity and atmospheric pressure also affect performance; high humidity can reduce sensor accuracy, while pressure fluctuations influence drone aerodynamics and disrupt the calibration of the GNC system \cite{rajawat2021weather,johnson2024lift}. Challenging weather conditions, such as rain and snow, create additional challenges, such as higher drag, lower propeller efficiency, and the potential for water damage to exposed electronic components \cite{siddique2022development}. To address these issues, medical drones are typically equipped with protective coatings and weather-resistant designs, enhancing reliability in diverse operational environments. Understanding these environmental influences is essential for selecting UAS platforms and developing robust GNC systems that ensure safe and efficient medical delivery missions.

\section{Payload Container Design}
The design of the payload container for medical drones must adhere to several essential constraints to guarantee efficiency and safety \cite{saunders2024autonomous}. First, the container needs to be lightweight and compact to reduce the extra load on the drone while also decreasing energy consumption during flight. It must also meet safety standards, ensuring dust-tight sealing and protection against strong water jets, thus shielding the container and its contents from mechanical impacts and external forces. Aerodynamic efficiency is another vital consideration as it lessens the power necessary for flight. Additionally, the choice of materials is crucial to the container's performance, with lightweight yet sturdy materials selected to preserve structural integrity while preventing unnecessary energy use. Finally, the transport container must be operationally compatible, providing flexibility in deployment across various aerial transport platforms \cite{vachalek2025design}.

\begin{figurehere}
\includegraphics[width=3in]{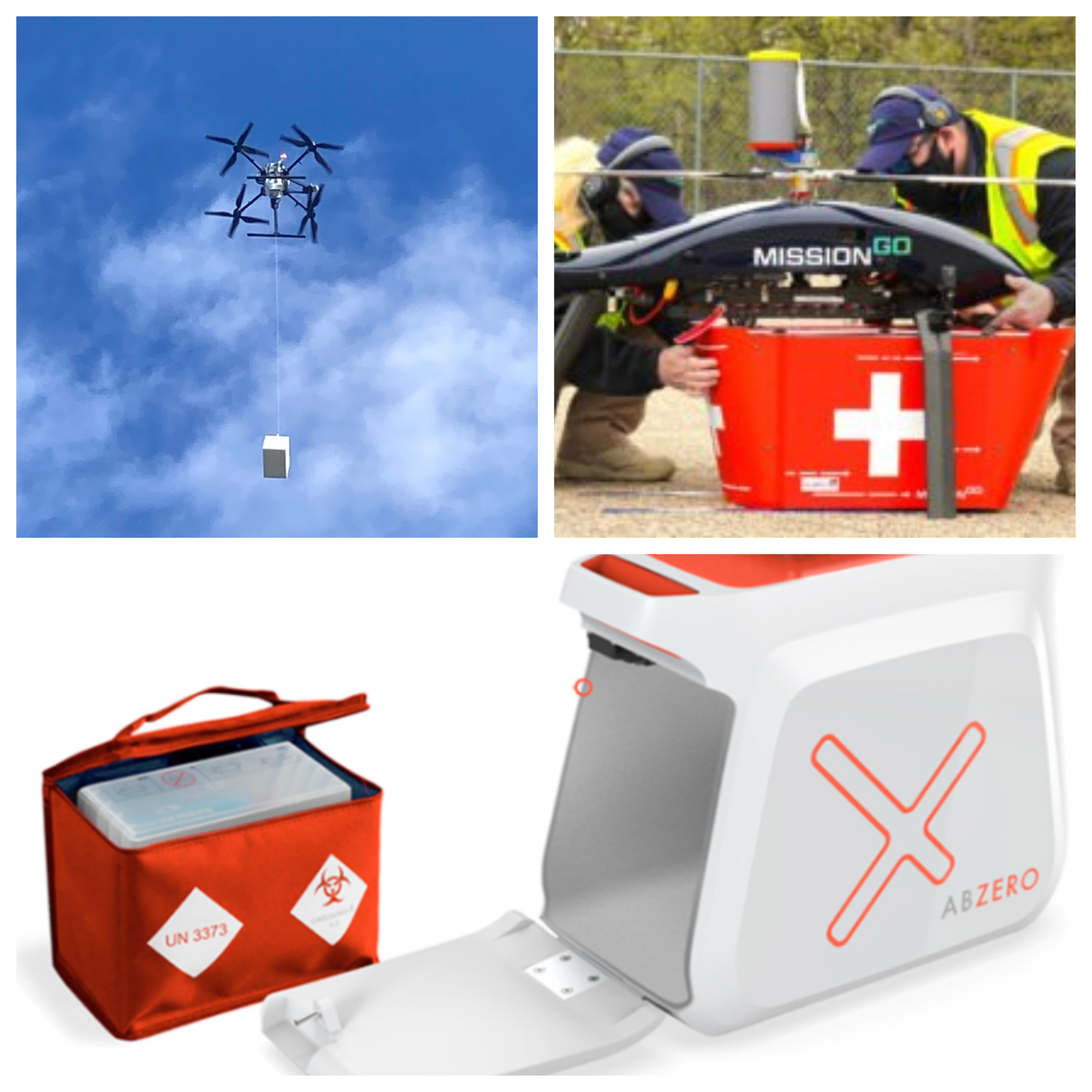}
\caption{Cable-Suspended \cite{mcnabb2025}, Fixed \cite{McNabb2021} and Temperature-Controlled Payload System\cite{amicone2021smart}.}
\label{fig:2}
\end{figurehere}

\subsection{Cable-Suspended Payload}
A cable-suspended payload system consists of a container that is suspended from the drone by means of a cable, enabling greater flexibility in payload and less aerodynamic disruption to the drone’s airframe. This method is especially beneficial for moving larger medical supplies or delivering items in difficult settings, like densely populated urban areas, disaster zones, or places where landing is impossible \cite{chen2019aerial}. The principal benefit of this system is that it permits drones to deliver packages without needing to land, which decreases the risk of contamination while transporting infectious materials such as COVID-19 test samples, vaccines, or dangerous biological agents \cite{mohsan2022role}. The construction of a cable-suspended system must take into account elements like stability, oscillation management, and effective payload release. The UAV with a cable-suspended load can be treated as a multi-body dynamical system \cite{omar2022hardware}. The equations of motion of each body can be formulated individually and then modified by adding the interaction forces between them using the Newtonian approach \cite{el2017nonlinear} or the Euler-Lagrange equations \cite{pizetta2015modelling}. The choice of cable length is vital, since the natural frequency of the system is inversely proportional to the square root of cable length, which in turn affects the vibration behavior of the payload, indicating that increasing the cable length will reduce the vibration in the system \cite{alkomy2021vibration}. To mitigate oscillations, drones utilize active damping systems and real-time control algorithms to modify flight dynamics and stabilize the suspended load \cite{geronel2022effect}.

Another important design factor is the location of the attachment point. Situating the suspension point close to the drone’s center of mass reduces the effects of payload movement on overall stability \cite{zeng2019geometric}. There are certain tasks that can only be performed by multiple drones, synchronized together to perform tasks such as carrying a load heavier than the payload capacity of a single quadrotor and successfully deliver an object under a rotor failure \cite{villa2020survey}.  Although cable-suspended payloads offer benefits such as enhanced delivery flexibility and the ability to release loads mid-air, they also present challenges in flight. The swaying motion of the suspended container can greatly influence stability, particularly in windy situations, which raises power consumption due to the frequent adjustments needed to maintain control \cite{bernard2011autonomous}. Additionally, the vibration of a cable suspended payload is a major challenge that requires a lot of challenge to solve \cite{villa2020survey}.  

\subsection{Fixed Payload}
A fixed payload denotes a container that is securely attached to the drone's airframe, ensuring improved stability and consistent flight performance as there are no additional degrees of freedom related to payload \cite{costantini2024cooperative}. This setup is commonly utilized because of its practicality and simplicity, although it is not suitable for heavy and large size packages and in some situations, it may not be possible to land the vehicle \cite{omar2023recent}. The design of a fixed payload system must take into account various factors to enhance performance. Adequate weight distribution is crucial to sustain the drone's center of gravity, since an imbalance can adversely affect maneuverability and control during flight \cite{pavithran2020prototype}. Furthermore, the structural integrity of the payload attachment needs to be strong enough to endure vibrations, wind disturbances, and flight movements, ensuring the container stays securely fixed throughout the operation \cite{estevez2024review}. Aerodynamic considerations also hold significant importance in the design of fixed payloads, as the container's shape directly influences drag and energy efficiency. A streamlined design lessens air resistance and improves flight endurance, which is especially advantageous for long-distance medical deliveries. Another vital element is vibration isolation since delicate medical supplies like vials and syringes must be safeguarded from mechanical shocks. This can be accomplished through the use of damping materials or a suspension system incorporated into the payload mount. The modularity of the design further enhances operational flexibility, permitting rapid payload changes for various medical applications. However, in spite of its benefits, the fixed payload system has drawbacks, such as limited cargo capacity and the necessity to land at the delivery location, rendering it less suitable for high-risk or difficult-to-access areas.

\subsection{Temperature-Controlled System}
The transport of medical supplies that are sensitive to temperature, like vaccines, blood samples, and biologics, demands a meticulously designed temperature-controlled system to maintain their integrity during transit \cite{gan2021apps}. Limited research has been conducted on the design of temperature-controlled container systems, with only a few studies available as references for developing an optimal design.

In \cite{amukele2017drone}, DJI S900 with a 4.7-litre cooler securely attached was used to transport blood products. The cooler contained pre-equilibrated thermal packs for platelets (20 to 24°C), wet ice for red blood cells (1 to 6 °C) and dry ice for frozen plasma (below -18°C) to maintain stable temperatures. A Type K thermocouple with a digital temperature logger monitored temperature stability, ensuring that blood samples remained within their required ranges despite ambient temperatures ranging from 2°C to 18°C. The properties of the blood didn't change after the flight. Likewise, in \cite{yakushiji2020quality}, drone transportation had no significant impact on the quality of red blood cell (RBC) solutions, as indicated by lactate dehydrogenase (LD) levels, which remained within acceptable limits post-flight. Transport was carried out using an M1000 quadcopter, and to maintain the required temperature range of 2 ° C to 6 ° C, an Active Transport Refrigerator (ATR 705) was used along with solid coolants that changed to a liquid state at approximately 4 ° C. While the internal temperature ranged from 4.7°C to 9.8°C, slightly exceeding the recommended range, the deviation was within a safe threshold for transfusion, confirming the feasibility of UAVs for blood transport.

Insulin is one of the most important lifesaving drugs for treating type I diabetes \cite{alemu2009insulin}. It is a sensitive peptide-based drug and is prone to irreversible aggregation when exposed to environmental stresses such as high temperatures and vibration, which can compromise its biological efficacy \cite{akbarian2018chemical}. In \cite{hiievaluation}, insulin was transported by securing vials in their original packaging on a DJI Mavic Air drone, flying them for approximately 9 minutes over 0.81 km, and analyzing post-flight stability. The environmental conditions during flight ranged from -1 to 0°C with 93 percent humidity, 2 mph wind speed, and air pressure of 1021 to 1022 mbar, while the vibration frequencies peaked at 3.4 Hz during take-off and remained at 0.1 Hz in flight. Active sensors or onboard environmental controls were not used, but post-flight tests confirmed that insulin quality remained unaffected. 

Likewise, \cite{costantini2024cooperative} presents the development of an insulated container, featuring a Peltier cooling system, Arduino-controlled temperature regulation, and thermal insulation to maintain a 2°C to 8°C temperature range. The system operates independently of the drone's power, using radiators, fans, and real-time sensors, with recommendations for enhancing energy efficiency and thermal stability using vacuum cavities and refrigerants. 

In \cite{amicone2021smart}, Smart Capsule was proposed, which is an AI-powered medical transport system designed for autonomous drone delivery, ensuring real-time monitoring and control of temperature, humidity, and agitation. It maintains optimal storage conditions using HDPE temperature stabilizers pre-cooled to 4°C, 22°C, or -25°C, providing up to three hours of stable temperature. Integrated with the spoke platform, it allows medical staff to manage deliveries without drone piloting skills. The system features GPS/GSM/4G connectivity for remote monitoring, a redundant GNSS positioning system for precise navigation, and an independent power supply for tracking in case of drone failure.

These research shows that a temperature-controlled payload system must ensure stable environment conditions. Key characteristics include precise temperature regulations using active cooling systems (e.g., Peltier elements, active refrigeration or phase-change materials) or passive methods (e.g., thermal packs, dry ice, or vacuum insulation). Real-time monitoring is crucial, often using data loggers, temperature sensors and remote communication (GPS/GSM/4G) to track conditions throughout transport. Systems could be independent of the drone's power so that it prevents failure and also maintains the temperature stability for extended periods. 

\section{Guidance Systems for Medical UAS}

\subsection{Minimum-Vibration Path Planning}
Designing dynamically feasible trajectories and controllers that drive a drone to a desired state with minimum vibration is a subject of interest for researchers and hobbyists. In trajectory generation, there are mainly two types of approaches. One of the most well-known probabilistic approaches is Rapidly-Exploring Random Trees (RRT) \cite{lavalle2001rapidly}, and many variations have been proposed, including LQR trees \cite{tedrake2009lqr} and RRT* \cite{karaman2011sampling}. These methods explore the space randomly while considering system dynamics to generate viable paths. Deterministic methods aim to compute optimal trajectories. Since the drone exhibits differential flatness, its trajectory planning can be effectively performed in a lower-dimensional flat output space. By solving a constrained Quadratic program (QP), minimum snap trajectories are generated, ensuring smooth and precise maneuvering. This approach facilitates the direct calculation of feasible paths while utilizing algebraic expressions instead of complex dynamic optimizations  \cite{mellinger2011minimum,mellinger2012trajectory}. Similarly, optimal polynomial trajectories can also be developed using a QP. However, the problem could be reformulated as an unconstrained QP, which enables numerical stability and facilitates the efficient optimization of long and complex trajectories while maintaining dynamic feasibility and collision avoidance. This method takes advantage of differential flatness \cite{murray1995differential} to eliminate the need for iterative simulations, greatly improving computational speed, and introduces a time allocation strategy that adjusts segment velocities based on a single aggressiveness parameter to achieve an optimal balance between smooth motion and fast traversal \cite{richter2016polynomial}. These methods have been successful in controlling the drone with a cable-suspended load, as the oscillation and vibrations of both the drone and load can be minimized using differential flatness and minimum snap (fourth derivative of position) optimization for smooth motion. \cite{sreenath2013trajectory}.

\subsection{BVLOS Trajectory Planning}
Currently, UAV applications are limited to operating within the visual line of sight of pilots. \cite{ivancic2019flying}. However, for applications such as package delivery, they are expected to fly autonomously over a long distance, where there is no visual line of sight and the pilot cannot observe the drone. Therefore, the further step in drone technology is to advance drones so that they can fly beyond the visual line of sight (BVLOS) \cite{behjati20223d}. BVLOS trajectory planning involves creating an efficient and safe flight path for drones operating beyond visual line of sight, requiring robust algorithms to concurrently avoid traffic, obstacles, and geofences, while considering real-time constraints and changes \cite{balachandran2017path}. BVLOS missions enable service providers to carry out complex operations using data from on-board and external instruments. The operator receives real-time information about the position, altitude, speed, and direction of the drone and relevant parameters of its structure \cite{davies2018review}. 
A UAV operating BVLOS should be capable of dynamically revising its path planning strategy according to the environment. This is achieved through a ``Detect Sense and Avoid (DAA)" system, which guarantees collision avoidance and enhances situational awareness \cite{gonzalez2020autonomous}. In order to improve the planning efficiency of drones, many innovative methods have been explored, for example, multi-resolution maps \cite{gonzalez2020autonomous}, game theory, and bioinspired methods, such as swarm optimization algorithms and potential fields \cite{iacono2018path}. Recently, the integration of Artificial Intelligence approaches, such as Machine Learning and Deep Learning, has brought attention to drone intelligence and automation, providing UAV applications with greater robustness and accessibility \cite{nouacer2020towards}. Reinforcement learning (RL) algorithms combined with deep learning could be used effectively in navigation tasks in complex environments \cite{wang2019autonomous}. Current cellular networks, including 5G, along with satellite links, may support data exchange in BVLOS applications \cite{hosseini2019uav}. Recent advancements in autopilot systems emphasize more intelligent adaptive control technologies that can sustain a higher degree of autonomy, enabling UAVs to effectively manage various environmental uncertainties as well as on-board limitations \cite{aber2019small}. Furthermore, the increasing demand for advanced decision-making and cognitive capabilities in modern aviation requires computationally efficient solutions, and recent research has explored the use of artificial neural networks \cite{boutemedjet2019uav}, genetic algorithms \cite{mousavi2019use}, fuzzy logic-based control techniques \cite{xie2019fuzzy}, and swarm intelligence \cite{dasdemir2020flexible}.

Despite these advancements, BVLOS flight planning remains a regulatory challenge, as governing bodies such as the FAA and EASA impose strict requirements for risk management, safety protocols, and real-time monitoring systems \cite{wanner2024uav}

\subsection{Trajectory Planning under Loss of Communication}
Planning the movement of UAVs when communication is lost is a major challenge. It requires smart strategies that help the UAV continue its mission safely using onboard systems, prediction methods, and backup plans. Various studies have emphasized the importance of maintaining reliable communication in UAV networks by optimizing their position, movement, and connectivity. Ensuring continuous network connectivity, minimizing disruptions during reconfiguration, and adapting to changing environments are key challenges addressed. Various strategies have been explored to enhance communication efficiency, reduce energy consumption, and improve network stability in dynamic scenarios. In \cite{grotli2012path,grotli2012path2,grotli2012task}, a mixed integer linear programming (MILP) approach is proposed for UAV mission and trajectory planning while considering radio communication constraints. In \cite{burdakov2010optimal}, an optimization method is presented for placing UAVs as relay nodes to ensure communication in environments with obstacles. It uses spatial discretization and shortest path algorithms to minimize the number of hops and path cost while maintaining line-of-sight connectivity. In \cite{han2009optimization}, the location and movement of UAVs are optimized to enhance the connectivity of a wireless network by addressing four key types of network connectivity - global message connectivity, worst-case connectivity, network bisection connectivity, and k-connectivity-through formulated optimization problems. In \cite{dimarogonas2008decentralized}, a decentralized mobility control algorithm is introduced for an optimal end-to-end communication chain using a team of unmanned aircraft serving solely as communication relays. In \cite{le2012adaptive}, a method for deploying an ad-hoc wireless network among UAVs is introduced, utilizing an optimization-based approach that separates the problem into two components, one for position optimization, and the other on improving communication, both interconnected through Lagrange multipliers. Numerous studies have focused on reconfiguring communication topology or addressing related challenges. In \cite{navaravong2012optimizing}, the study explores the challenge of transitioning from an initial connected formation to a desired final connected formation while ensuring continuous connectivity throughout the reconfiguration process. In \cite{milner2012nature}, two biology-inspired techniques, a Flocking Algorithm (FA) and a Particle Swarm Optimization (PSO) algorithm, are employed to control a hierarchical heterogeneous wireless network to adapt to changing environments and reducing overall energy consumption. In \cite{spanos2005motion}, the challenge of preserving network connectivity during a reconfiguration maneuver is examined. A metric is introduced to assess the extent to which individual vehicles can move freely without causing network disconnection. 

\section{Navigation Systems for Medical UAS}

Navigation systems determine a UAV's state, including its position and velocity \cite{MARSHALL2021390}. In most real-world applications, the UAV's surroundings are unknown, making navigation challenging. This is crucial in medical applications, where payloads must be securely protected and safely delivered.

\subsection{Sensors}
Medical drones require high precision, reliability, and navigation autonomy because they operate under various, often challenging conditions. Drones use a variety of sensors that provide them with real-time information about their environment to accomplish these tasks, enabling obstacle identification, localization, and flight stability.

Different kinds of sensors are used for specific purposes, including positioning, object detection, environmental awareness, and collision avoidance. For navigation and target recognition, optical sensors (e.g., RGB and infrared cameras) capture visual information. Depth perception and terrain mapping are made possible by distance-measuring sensors, such as LiDAR, radar, and ultrasonic sensors. Positional sensors (e.g., GNSS and IMU) are essential for autonomous operation, as they ensure precise localization and flight stability.

\subsubsection{Optical Sensors}
Optical sensors, i.e., cameras, are one of the most popular sensors in medical drones because of their affordability, adaptability, and capacity to record high-resolution data. 

One of the most common cameras in UAV applications is the RGB camera, which captures images within the visible light spectrum. RGB cameras are widely employed for certain tasks such as object detection, scene interpretation, and real-time environmental monitoring. For example, they are used in obstacle detection and avoidance systems, where the UAV can identify and navigate around structures such as buildings by combining depth imagery with occupancy voxel maps \cite{10161194}. In addition to obstacle avoidance, RGB cameras are also utilized for terrain detection. This is typically achieved by enhancing the image data with a Digital Terrain Model (DTM), which provides topographic context crucial for ensuring safe and accurate deliveries in rural or uneven landscapes \cite{ijgi10050285}.

On the other hand, in low-visibility operations, such as those involving fog, smoke, or nighttime missions, where normal RGB cameras may perform poorly, infrared (IR) cameras become particularly useful. The capability to detect heat signatures rather than visible light makes them perfect for certain applications, such as autonomous precision landing, patient detection, and thermal monitoring. 

Common applications of IR cameras in medical drones are autonomous precision landing, which is critical when delivering medical supplies to hospitals, mobile emergency units, or remote locations. For example, the IR camera is equipped on a fixed-wing UAV in order to perform a night landing on a runway \cite{aerospace9100615}. Additionally, for multi-rotor UAVs, the landing process is enhanced by cooperating with an infrared camera and an infrared beacon placed on the landing pad \cite{9530091,8388325}. This ensures a successful delivery by enabling the drone to find the landing area in difficult circumstances. 

Additionally, IR cameras are useful in search-and-rescue (SAR) missions, where they help locate injured people based on body heat signatures \cite{drones5030065}. This feature is essential in disconnected or disaster-affected places where patients may be stuck in low-visibility conditions or trapped under debris.  

By integrating RGB and infrared cameras, medical drones can navigate more effectively and operate in diverse environmental conditions. These advancements ensure that life-saving supplies and medical interventions reach those in need, regardless of obstacles or environmental conditions.

\subsubsection{Distance-Measuring Sensors}
Distance-measuring sensors are another essential sensor that leads to the safe navigation of medical drones. These sensors allow UAVs to precisely determine the distances to obstacles, terrain, and other objects, which are crucial capabilities for precision landing \cite{10.1117/12.904062}, altitude control \cite{7856456}, and collision avoidance \cite{6306078, 1336460}. Unlike RGB and infrared cameras, distance measuring sensors measure precise depth perception, allowing medical drones to function more accurately in difficult surroundings. 

LiDAR (Light Detection and Ranging) sensors are the first example of sensors that measure distance. It measures how long it takes for light from laser pulses to bounce off nearby objects in order to calculate distances. LiDAR is widely used in autonomous vehicles \cite{s22010194}, including UAVs, where it assists with tasks such as power line inspection \cite{GUAN2021106987}, which is useful for navigation of an obstructed operation. LiDAR's ability to function regardless of terrain or flight conditions is one of its main advantages. This makes it perfect for missions in remote or difficult-to-reach areas \cite{Abazaj_Llabani_Qirjazi_2024}, as well as in crowded urban settings where traditional sensors might have trouble because of noise or obstructions \cite{rs14010238}. Another important feature of LiDAR is its capacity to generate high-resolution 3D maps \cite{9501167, s17102371}, which enhances the drone’s spatial awareness and obstacle avoidance skills. This is especially helpful for negotiating crowded locations, emergency response, and precise delivery. LiDAR is frequently used with other sensors, such as global navigation satellite systems (GNSS) and inertial measurement units (IMUs), to increase detection stability and accuracy \cite{10285475}. Even in situations where GPS is unavailable or the environment is visually complicated, this sensor fusion method offers more accurate positioning and mapping.

Though sound-based sensors work similarly to LiDAR, they use high-frequency sound waves rather than laser pulses to measure the time it takes for sound to bounce off an object \cite{Terzic2013}. The ultrasonic sensor is among the most widely utilized sensors in its class for UAV applications. Ultrasonic sensors are perfect for short-range sensing applications, since they are highly effective in detecting near objects, for example, the drone equipped with the ultrasonic sensor is proposed to be utilized for interior vessel inspection \cite{10032580} and indoor flight \cite{9440350, app11156805}. Moreover, they are used in UAVs to help with accurate and autonomous landing \cite{9651358, 9214041}. In medical drone applications, where drones would have to land on tiny platforms such as roofs or authorized landing zones in emergency locations, this is very important. Their ability to provide reliable altitude feedback at low altitudes helps ensure safe and accurate touchdowns. Their advantages include being low-cost, lightweight, and simple to integrate \cite{mi13040520}, which makes them suitable for small UAVs. They also perform well in visually cluttered environments where optical sensors may have trouble. However, their drawbacks include a short detection range (typically a few meters) and lower accuracy in noisy environments, such as strong winds or motor vibrations \cite{BUTT2024104629}. They may also be less accurate when detecting soft or angled surfaces that absorb or deflect sound waves. 

Radar sensors allow UAVs to identify obstacles, measure distances, and estimate object velocity—even in situations where other sensors would not work \cite{10.1007/978-3-030-71151-1_1} by emitting radio waves and timing how long it takes for them to bounce back off surrounding objects \cite{kingsley1999understanding}. For medical drone missions that need reliability in rain, fog, or dust, radar is particularly useful for long-range navigation. Radar allows beyond-visual-line-of-sight (BVLOS) operations in rural or obstructed terrain and offers early identification of distant impediments \cite{8571665}, in contrast to ultrasonic sensors, which are best suited for short-range tasks. However, radars usually have a poorer spatial resolution compared to LiDAR or cameras. They might also have trouble identifying small or complicated objects, and are more expensive and heavier \cite{drones6070167}, which can be a barrier for lightweight UAV platforms.

\subsubsection{Positional Sensors}
Positional sensors, which determine location, orientation, and movement, are crucial to the safe and precise navigation of medical drones. The Global Navigation Satellite System (GNSS) and Inertial Measurement Units (IMUs) are the two most widely utilized sensors in this category. Drones can establish routes, maintain steady positions, and perform precise movements \cite{Cahyadi2023, chi2020robust} due to these sensors, which form the basis of autonomous flying. This is very helpful when transporting medical supplies to remote or unpredictable destinations. 

GNSS, which consists of systems such as GPS, GLONASS, Galileo \cite{LUO20214916}, and BeiDou \cite{Chen_Zhao_Liu_Zhu_Liu_Yue_2021}, triangulates satellite signals to provide global coordinates \cite{ogaja2022introduction}. It enables precise geolocation monitoring, which becomes useful for long-distance tasks when the drone needs to adhere to a predetermined route. However, in some locations, such as crowded cities (because of signal interference) and underground environments, GNSS signals may be inaccurate or unavailable \cite{peng2024gnss}. Drones require backup or auxiliary navigation systems in these situations to ensure flying stability. 

IMUs measure the drone's acceleration, angular velocity, and magnetic field. They are composed of accelerometers, gyroscopes, and magnetometers \cite{ESSER20091578}. Depending on the grades (e.g, consumer, industrial, tactical, and navigation), IMUs can enable the drone to navigate itself and estimate position changes while offering it real-time feedback on motion and attitude even in situations when GNSS is unexpectedly unavailable \cite{s151026212}. IMUs support flight stability and reactivity in emergency \cite{meng2025mars} or dynamic situations, including flying through choppy weather or restricted rescue zones \cite{lyu2023survey}.
IMUs are frequently merged with GNSS data using techniques like the Kalman filter, which will be discussed in the next subsection, which produces a far more robust and dependable location estimate, even if they have a tendency to drift over time without additional correction. 

\subsection{Sensor Fusion Approaches}
Since medical drones often operate in uncertain, dynamic, or complex environments, such as during snowfall or rainfall \cite{doi:10.1177/03611981211036685} and urban environment \cite{10.1007/978-3-031-76597-1_52}, relying on a single sensor type can lead to inaccurate or incomplete perception. Sensor fusion is the process of combining data from multiple sensors to enhance the reliability, accuracy, and robustness of the navigation. 

Sensor-filtering fusion techniques, such as the Kalman Filter and Extended Kalman Filter (EKF), are among the most widely used in UAV systems. These methods estimate the drone’s state by recursively combining noisy measurements from different sensors, such as GNSS, IMUs, and barometers \cite{Cahyadi}, to produce a more accurate and smoothed position or velocity estimate. In medical drones, KF-based approaches are likely to be used to stabilize flight \cite{Navisa_2023}, especially when GNSS signals are weak or intermittent \cite{9836124, electronics8050478}.

Optimization-based fusion methods formulate the state estimation problem as an optimization task, where the objective is to minimize the error between the predicted and observed measurements over a window of time. This class includes methods like factor graphs \cite{doi:10.1177/0278364906072768, 4682731} and bundle adjustment \cite{10.1007/3-540-44480-7_21}, often used in Simultaneous Localization and Mapping (SLAM) frameworks.

With the rise of deep learning, learning-based fusion has gained popularity, especially in vision-based and high-dimensional sensor integration. These approaches leverage neural networks \cite{ZHOU20222878}, such as CNNs \cite{10175389} or RNNs \cite{ASGHARPOORGOLROUDBARI2023113105}, to learn the mapping between raw sensor data (e.g., images, LiDAR, IMU) and navigation outputs. For medical drones, learning-based fusion is particularly useful for tasks like visual-inertial odometry, gesture recognition, or adaptive obstacle detection, where conventional filtering may struggle due to highly dynamic or cluttered scenes. Additionally, machine learning can enable adaptive sensor weighting, allowing the system to prioritize more reliable sensors based on the environment or context.

\subsection{Environmental and Situation Awareness}
Environmental and situational awareness refers to a UAV’s ability to perceive, interpret, and model its operational surroundings \cite{10749484,10.2514/6.2025-2436}. This capability is particularly vital for medical drones, which often operate in uncertain, unstructured, or rapidly changing environments, where both flight safety and mission success depend on accurate environmental understanding. Relevant environmental factors include static obstacles such as buildings, trees, and terrain; dynamic elements such as vehicles, pedestrians, and debris; and variable weather conditions like rain, fog, and dust.

To construct an understanding of these factors, UAVs rely on a suite of onboard sensors, including RGB and thermal cameras, LiDAR, radar, and ultrasonic sensors, along with data processing techniques that integrate raw inputs into a coherent environmental model. For example, LiDAR has been extensively used for indoor mapping and obstacle detection \cite{HU2024108901, 10148919}, which can be utilized with outdoor objects, while thermal cameras have proven effective in detecting humans under low-visibility conditions \cite{https://doi.org/10.1002/rob.21985}.

Enhancing this awareness involves applying a range of algorithmic methods. Optimization-based techniques allow for the identification and prediction of dynamic elements like vehicles \cite{9430769} and pedestrians \cite{9374712} by minimizing discrepancies between predictions and observations. In addition, where conventional sensing is limited due to infrastructure loss or communication constraints, deep learning methods, particularly Convolutional Neural Networks (CNNs), enable the detection of aerial targets \cite{10183872} and classification of static structures using labelled datasets \cite{app112411611}. Another learning-based approach is the reinforcement learning (RL) method. It has been applied to infer missing environmental information and enable adaptive scene interpretation \cite{10041962, drones7020123}.

This real-time environmental perception forms the basis for higher-level decision-making systems, including obstacle avoidance, trajectory planning, and safety-critical flight control, which are discussed in the following section.

\subsection{Detect-and-Avoid Approaches}
While environmental awareness provides a comprehensive understanding of the UAV's surroundings, detect-and-avoid (DAA) systems are responsible for translating that understanding into real-time safety actions. These systems enable UAVs to anticipate and respond proactively to potential hazards, modifying their flight paths to prevent collisions with static and dynamic obstacles.

A widely used approach relies on 3D sensing technologies, such as LiDAR, to generate spatial maps that allow UAVs to detect nearby objects and adjust their trajectories accordingly \cite{drones6080185, drones6010016,perumalla2024passive}. For instance, SLAM-integrated 3D sensors have been shown to support real-time collision avoidance in static indoor environments.

In \cite{9476746}, the author demonstrates the use of monocular vision—a vision-based sensing modality—to achieve the detect-and-avoid function, showcasing its potential for lightweight UAV platforms operating in constrained environments.

In \cite{Zuo_2021}, various sensor configurations for obstacle avoidance are evaluated, highlighting the trade-offs among cost, robustness, and precision between RGB-D cameras, LADAR systems, and camera–IMU combinations. The results show that RGB-D cameras underperform in complex or low-visibility environments. In contrast, LADAR systems offer higher precision and environmental robustness, albeit with significantly greater power consumption. Monocular and binocular camera–IMU combinations are explored as lightweight alternatives, with the binocular setup providing improved accuracy at the cost of increased computational complexity.

In terms of algorithmic response, learning-based approaches such as reinforcement learning have gained traction for adaptive obstacle avoidance. For example, the RL framework in \cite{9537945} allows UAVs to learn dynamic avoidance policies, while \cite{drones7020123} extends this to handle both static and moving obstacles in real time. In more complex deployments, deep reinforcement learning techniques like dueling double deep Q-networks (D3QN) have been applied to path planning in non-cooperative multi-agent settings, optimizing both collision avoidance and data collection from distributed nodes \cite{9718529}.

Likewise, bio-inspired optimization algorithms, such as the ant colony optimization method with variable pheromone tuning (ACO-VP) \cite{10130088}, have been used to plan collision-free paths in dynamic coverage tasks, demonstrating efficiency improvements in target reassignment and route reconfiguration.

Additional methods focus on global coordination goals. For example, quality-of-coverage (QoC) optimization \cite{8701486} and multi-agent deployment strategies \cite{9606204} aim to enhance spatial distribution while minimizing interference and collision risk.

Together, these DAA strategies leverage perception, prediction, and planning to ensure safe and adaptive operation, particularly critical for medical drones tasked with navigating through uncertain or congested airspace.

\section{Control Systems for Medical UAS}

\subsection{Vibration Control}
UAS used for medical deliveries are susceptible to vibrations, which can affect flight stability and the integrity of sensitive medical payloads. Vibration control in UAVs is crucial for medical deliveries as they experience several types of vibrations, such as vibrations from engines, propellers, aerodynamics, and structural components. Vibrations can damage UAV components like sensors, batteries, and electronics, reducing performance. Likewise, prolonged vibrations may cause structural fatigue, leading to cracks and weakening stability. Intense vibrations can also disrupt flight control, causing instability or loss of control \cite{li2025investigation}.

Vibration reduction technologies are classified as active or passive. Active control systems detect vibrations and respond to suppress them, offering better performance but requiring complex sensors, algorithms, and actuators, increasing costs. Passive systems, however, use damping materials to absorb or isolate vibrations without external control or energy, making them simpler and more cost-effective \cite{zeqi2017some}. Passive vibration control systems are known for their reliability, simplicity, and cost-effectiveness, making them suitable for a wide range of engineering applications. Rubber dampers, non-angular displacement dampers, and metal dampers are some of the passive vibration control systems \cite{jiang2021integrated, balaji2021applications}. 
Research on vibration reduction for UAV-mounted cameras explores passive and active methods. Passive approaches include magnetic isolators \cite{kienholz1996magnetically}, friction dampers \cite{gjika1999rigid}, and viscoelastic materials \cite{webster2005broad} for isolation. Wire rope isolators \cite{maes2014passive} and active systems like voice coil actuators \cite{verma2020active} enhance vibration control. These studies show that combining advanced isolators and frequency-dependent materials improves performance. Likewise, vibration control approaches include rubber dampers for lidar systems \cite{changshuai2019vibration,fu2022design}, foam to reduce accelerometer errors \cite{maj2009novel}, and vibration isolation platforms using steel springs and carbon fiber panels \cite{brown2022vibration}. Structural vibration studies on multirotor UAVs highlight that while these technologies enhance performance, they can complicate the structure, increase failure risk, and consume more energy, reducing flight duration \cite{chen2023investigation}. 

In recent years, mechanical metamaterials have gained attention due to their unique properties that surpass those of their base materials \cite{liu2000locally,smith2004metamaterials}. These metamaterials, often composed of periodically arranged components, exhibit characteristics such as negative Poisson's ratio, bistability, nonlinearity, and tunable stiffness, making them ideal for vibration control and sound wave propagation \cite{ji2021vibration}. Common types include cellular metamaterials \cite{gramuller2014pacs}, bandgap metamaterials \cite{reynolds2017enhancing}, auxetic metamaterials with negative Poisson's ratio \cite{buckmann2012tailored}, and Pentamode metamaterials with significantly higher bulk modulus compared to shear modulus \cite{kadic2013anisotropic}. They offer various applications, such as low-frequency noise absorption using thin-film acoustic metamaterials \cite{mei2012dark}, wave propagation control through waveguide piezoelectric metamaterials \cite{del2014dynamic}, and dynamic energy dissipation by concave hexagonal honeycombs \cite{fu2016nonlinear}. Additionally, nanolattice mechanical metamaterials demonstrate ultra-high strength and stiffness, making them suitable for energy absorption \cite{bauer2017nanolattices}. 

Research indicates that some medicinal products, such as proteins in biopharmaceuticals and blood, may undergo aggregation when subjected to vibration \cite{hiievaluation,torisu2017friability}. However, limited research has explored the consequences of transport-related vibration on the quality of medical products. Specifically, vibrations generated from drone platforms pose an uncertain risk to product quality, as they typically occur at higher amplitudes and frequencies compared to conventional road transport methods such as cars and bicycles \cite{oakey2021quantifying}. Addressing these vibrations is essential to preserving the efficacy of medical goods transported. Practical tests comparing a twin four-stroke fixed-wing drone (1100-3400 RPM) and a battery-powered hexacopter (2000-2500 RPM) showed that the hexacopter produced higher vibrations. Both drones generated vibrations at significantly higher frequencies than typical road vehicles (usually under 20 Hz). The trials involved flying Actrapid insulin to examine if its quality is affected by vibrations. Laboratory tests confirmed that all insulin samples passed the British Pharmacopoeia turbidity test, indicating no loss in quality. However, the study highlighted two key findings: (I) It's crucial to understand how sensitive medicines are to different vibration frequencies before transporting them by drone, and (ii) The standard medical packaging used did not effectively protect  Actrapid from drone-induced vibrations, especially in the fixed-wing drone \cite{oakey2021quantifying}. In \cite{waters2024design}, researchers designed packages equipped with coil springs and wire rope isolators, comparing their performance to a standard National Health Service (NHS) medical package through lab and real-world drone and road transport tests. Both prototypes effectively minimized the high-frequency vibrations commonly produced by drones, reducing them by up to six times. However, they were slightly less effective during road transport due to the lower-frequency vibration environment. These findings highlight the potential of customized packaging to protect the quality of the medicine during drone delivery. In addition, certain flight phases, such as take-off and landing, exhibit higher vibration levels \cite{mansfield2022whole}. This underscores the necessity for enhanced vibration control measures, particularly during critical flight segments, to maintain the stability of transported medical products. 

\subsection{Thermal Control}
The operational efficiency and reliability of UAV systems are significantly influenced by thermal conditions, which affect not only the onboard electronic components but also the integrity of sensitive payloads. Ensuring effective thermal control is essential to maintain the optimal performance of medical UAS. Critical electronic components, including gyroscopes, crystals, GPS, and other electronic modules that are highly sensitive to temperature changes. When the internal temperature fluctuates significantly, these variations can cause substantial errors due to temperature drift, compromising the system's operational reliability \cite{li2023temperature}. Additionally, in low-temperature environments, the discharge capacity of lithium polymer batteries used in UAVs decreases significantly \cite{piao2022challenges}. When the temperature falls to -20°C, the battery may enter hibernation mode, preventing it from functioning properly.

For medical payloads, maintaining specific temperature ranges is also crucial to preserving the efficacy of temperature-sensitive medicines and vaccines. Exposure to temperatures outside recommended storage conditions can degrade these medical products, rendering them ineffective or even harmful. Therefore, precise thermal management during UAV flights is important to ensure the safe delivery of medical supplies. To address these challenges, both passive and active thermal control strategies are employed. Passive methods include the use of thermal insulation materials and phase change materials to maintain stable internal temperatures by absorbing and releasing heat in response to external temperature changes. Active thermal control involves integrating thermoelectric coolers and heating elements that dynamically regulate the internal environments of the UAV, ensuring that both electronic components and medical payloads remain within their optimal temperature ranges.

PID control is a classic control method that is composed of proportional, integral, and differential negative feedback systems. They are commonly used in thermal management systems but often require precise tuning and may not adapt well to varying environmental conditions \cite{sanguino2024design}. Likewise, Fuzzy control is extensively applied in various automated systems, including temperature, flow, and pressure regulation \cite{feng2006survey}. It is a control technique based on fuzzy logic, primarily used for systems that are difficult to define and model with precision. Fuzzy PID controllers enhance traditional PID systems by incorporating fuzzy logic, allowing real-time adjustment of control parameters based on the system's behavior and external conditions. This adaptability improves the system's ability to maintain thermal stability under fluctuating temperatures, which is particularly beneficial for uAV operating in diverse environments \cite{phu2020new} \cite{kumar2011review}.  

\subsection{Icing Control}
Adverse weather is one of the key factors that can hinder the performance of UAVs \cite{muller2023uav}. A common adverse weather challenge UAVs encounter is atmospheric icing \cite{cao2018aircraft}. When a UAV flies through a cold-containing supercooled droplet, the droplets that hit its surface freeze instantly upon contact, which will have a negative impact on the aerodynamic performance of the UAV \cite{szilder2017flight}. Ice accumulates on the airfoil, altering its aerodynamic shape, which reduces the ability of the wings to generate lift and increases the aircraft's drag \cite{bragg2005iced}. Research on UAV icing has only recently gained momentum, with a significant portion concentrating on icing effects on UAV airfoils \cite{hann2020unsettled}. Because of their typically smaller size and high relative air speeds (due to rotation), propellers are more susceptible to icing than wings \cite{muller2021uav}. 

In icing wind tunnel experiments, it has been shown that ice accretion on propellers will reduce the propeller's thrust and increase the power needed to operate them \cite{liu2018experimental} \cite{liu2019experimental}. This rapid thrust reduction can quickly cause a loss of control of the UAV, particularly in multi-rotor UAVs that rely on the rotor thrust for both lift as well as control \cite{yan2020experimental}. A unique challenge associated with icing on rotating parts such as propellers and rotors is the centrifugal force generated during rotation, which can lead to ice shedding \cite{nilamdeen2019ice}. Ice shedding refers to the phenomenon in which accumulated ice detaches from the propeller because the adhesive force between the ice and the propeller surface is insufficient to hold the ice in place. Although this process may recover some of the propeller's lost performance, it can also introduce potential hazards to the UAV \cite{muller2022uav}. If ice detaches from only one blade, it creates a mass imbalance in the propeller, leading to significant vibrations throughout the system. In addition, the aerodynamic forces become uneven across the blades, further increasing the mechanical stress. In addition, detached ice fragments can strike other components of the UAV, potentially causing structural damage. 

A common solution to ice buildup on UAV propellers and rotors is the use of ice protection systems (IPS). This system functions by either preventing ice formation (anti-icing) or periodically removing it after limited accumulation (de-icing). A major challenge for UAVs is limited power availability. Suitable IPS options for UAVs include special surface coatings and electro-thermal systems, which convert electrical energy into heat to prevent icing \cite{thomas1996aircraft}. In \cite{liu2018finite}, the impact of superhydrophobic coatings on propellers was analysed. Compared to uncoated propellers, coated ones showed a 70 percent decrease in additional power consumption at -5°C and a 75 percent reduction in thrust loss. These improvements were linked to reduced ice accumulation outside the impingement zone and faster ice shedding. The research was done on the performance of several commercially available coatings on a scaled small UVA rotor. At -12°C, only one coating effectively reduced ice accumulation to a level that allowed the UAV to sustain hovering. However, it resulted in a  28 percent drop in thrust and a 50 percent rise in torque \cite{villeneuve2022experimental}. In \cite{laroche2021silicone}, a method was introduced for metallic alloys to achieve durable hydrophobic and ice-phobic characteristics. Building on this, \cite{alamri2020self} proposed that incorporating these surfaces could enhance the performance of thermal ice protection systems. In \cite{karpen2022propeller}, a rotor with an integrated ice protection system was developed for multi-rotor UAVs. The propeller was coated with a carbon-based heater paint, divided into three zones, each with a fixed heat output determined by its electrical resistance. This paint offers passive temperature control due to its strong temperature response at 60°C. A supercapacitor in the rotor hub powered the system, allowing it to prevent icing for 15 minutes in glaze ice at 0°C. Although it couldn't fully prevent icing in rime conditions at -10°C, it did increase ice-shedding frequency, reducing overall icing effects. Developing an IPS for rotating parts like propellers is more difficult than for wings. This is because the heat required changes along the length of the propeller due to changing airflow \cite{samad2021experimental}. Additionally, because propellers are smaller than wings, it's harder to fit heating elements and maintain precise assembly. Another issue is how to deliver power to the rotating propeller. One solution is to use a power source inside the rotor hub \cite{karpen2022propeller}, or to send power through the motor from the UAV, as done by \cite{fengler2017study} in their electro-thermal IPS design.

\section{Discussion}
The findings synthesized in this survey underscore a critical theme: the success of a medical drone mission is contingent on a deeply integrated approach to its design and operation, where GNC, flight dynamics, and payload integrity are inextricably linked. The review of UAS configurations reveals a fundamental trade-off between the aerodynamic efficiency and range of fixed-wing UAVs and the operational flexibility of multirotor UAVs, which offer VTOL and hover capabilities ideal for urban or confined areas. The emergence of hybrid VTOL designs attempts to balance these trade-offs, making them particularly suitable for reaching remote regions. This choice of airframe directly informs payload integration. A fixed payload offers greater stability by avoiding additional degrees of freedom, but requires the drone to land, which may be unfeasible in disaster zones. Conversely, a cable-suspended payload allows for package delivery without landing, reducing contamination risk, but introduces significant dynamic challenges, such as oscillation, that demand advanced control algorithms to manage.

Interpreting these findings, it becomes clear that environmental factors are a dominant and pervasive challenge that impacts every facet of the mission. For instance, temperature extremes not only degrade battery performance, shortening flight times, but also directly threaten the viability of temperature-sensitive payloads like blood products and vaccines. Similarly, vibration, whether from propellers or aerodynamic forces, poses a dual threat: it can degrade sensor performance, affecting flight control, and compromise the molecular integrity of sensitive biologics like insulin through aggregation. This survey highlights that standard medical packaging is often insufficient to protect against the high-frequency vibrations typical of drones. The implication is that advanced, mission-specific packaging with passive or active damping is not an accessory, but a core component of the GNC system. Furthermore, atmospheric icing presents a severe hazard, as ice accretion on propellers can rapidly lead to a loss of thrust and control, a critical failure point for multi-rotor systems.

The GNC algorithms and systems reviewed in this paper should be viewed as direct responses to these multifaceted challenges. Minimum-vibration path planning, using techniques like differential flatness and minimum snap, is a guidance-level solution to preserving payload quality. Robust navigation, achieved through the fusion of multiple sensors like GNSS, IMU, and LiDAR via Kalman filters , is essential for maintaining accuracy and safety, especially when operating BVLOS or in GNSS-denied environments. The broader implication of these findings is that designing medical drones requires a holistic systems-engineering perspective. An off-the-shelf drone is rarely adequate; rather, the airframe, payload container, and GNC architecture must be co-designed to withstand anticipated environmental stressors while guaranteeing the safety and efficacy of the medical cargo.

Future research directions should focus on several key areas. First, the exploration of advanced materials, such as mechanical metamaterials, for passive vibration isolation offers a promising, low-power solution to protecting sensitive payloads. Second, the development of intelligent payload systems—containers with integrated, real-time monitoring of temperature and vibration that can provide feedback to the GNC system to alter flight paths or speeds—could create a fully adaptive delivery platform. Third, while AI and reinforcement learning are being used for path planning and obstacle avoidance, future work could extend these techniques to holistic mission management, where the AI controller optimizes for flight time, energy consumption, and payload integrity simultaneously. Finally, developing lightweight, low-power, and effective ice protection systems (IPS) for propellers remains a significant challenge and is a critical area of research to enable reliable all-weather operations.

\section{Conclusions}
This paper has presented a comprehensive survey of medical drones, with a specific focus on the challenges and solutions from the perspectives of flight dynamics, guidance, navigation, and control. The review consolidates critical insights into the technical requirements for designing and deploying UAS for safe and effective healthcare delivery.

The principal conclusions drawn from this survey are as follows:

\begin{itemize}
    \item The selection of a UAS configuration, whether fixed-wing, multirotor, or hybrid VTOL, is a foundational design choice that creates a trade-off between range, endurance, and operational flexibility, and must be aligned with specific medical mission requirements.
    \item Payload container design is an integral component of the overall system, not a passive attachment. Its design, whether fixed, cable-suspended, or temperature-controlled, directly impacts flight stability and, most importantly, the viability of the medical supplies being transported.
    \item Environmental conditions, including wind, turbulence, temperature, and atmospheric icing, pose significant risks to flight performance, electronic reliability, and payload integrity, necessitating the development of robust and adaptive control systems.
    \item Advanced GNC systems are paramount for mission success. This includes guidance strategies for minimum-vibration path planning, sophisticated sensor fusion techniques for reliable navigation in challenging environments, and robust detect-and-avoid systems to ensure safety during autonomous and BVLOS operations.
    Ultimately, the successful deployment of medical drones at scale depends on the synergistic integration of these GNC technologies. Addressing the challenges of environmental variability and payload sensitivity through advanced control and design is essential to realizing the full potential of UAS to transform modern healthcare logistics.
\end{itemize}

\bibliographystyle{ws-us}
\bibliography{reference1,conferences1,journals1,patents1}

\noindent\includegraphics[width=1in]{images/author1.pdf}
{\bf Roshan Kumar Chhetri}  received his Bachelor’s degree in Mechanical Engineering from Kathmandu University, Nepal. He is currently pursuing his M.S. degree in Mechanical Engineering at Texas Tech University, Lubbock, USA. His research interests include control, optimization, robotics, and unmanned aerial vehicle dynamics. 

\noindent\includegraphics[width=1in]{images/author2.pdf}
{\bf Sarocha Jetawatthana }  is a graduate research and Ph.D student in the department of Mechanical Engineering, Texas Tech University. She received her B.Eng. Degree in Aerospace Engineering from Kasetsart University,  Thailand. Her research focuses on state estimation in the guidance, navigation, and control of autonomous system and unmanned aerial vehicle dynamics.

\noindent\includegraphics[width=1in]{images/author3.pdf}
{\bf Dr. Thanakorn Khamvilai} received his M.S. degree in Aerospace Engineering from Georgia Institute of Technology in 2017 and his Ph.D. degree in Aerospace Engineering (with a minor in Electrical and Computer Engineering) from Georgia Tech in 2021. He served as Assistant Research Professor at the Pennsylvania State University Unmanned Aircraft Systems Research Laboratory, working on real-time optimization and autonomous systems guidance, navigation, and control. He is currently Assistant Professor in the Department of Mechanical Engineering at Texas Tech University. Dr. Khamvilai’s research interests include aircraft guidance, navigation and control, avionics systems and aviation, and optimization. He is an active member of the American Institute of Aeronautics and Astronautics (AIAA) Digital Avionics Technical Committee.

\end{multicols}

\end{document}